\pdfoutput=1
\documentclass[11pt]{article}
\usepackage[final]{acl}

\usepackage{times}
\usepackage{latexsym}
\usepackage[T1]{fontenc}
\usepackage[utf8]{inputenc}
\usepackage{microtype}
\usepackage{graphicx}
\usepackage{booktabs}
\usepackage{array}
\usepackage{placeins}
\usepackage{amsmath,amssymb}

\title{Equal Difficulty, Unequal Confidence:\\
LLM Calibration and Abstention in African Languages}

\author{St John Grimbly \\
  Department of Mathematics and Applied Mathematics \\
  University of Cape Town \\
  \texttt{uct@stjohngrimbly.com}}

\hypersetup{pdftitle={Equal Difficulty, Unequal Confidence: LLM Calibration and Abstention in African Languages},pdfauthor={St John Grimbly}}

\begin{document}
\maketitle

\begin{abstract}
Evaluations of large language models in African languages rarely test whether a model's
confidence can guide deferral to a human, and cross-lingual confidence comparisons seldom adjust
for question difficulty. We study three confidence signals from four model families
($1.5$--$32$B) on five multiple-choice benchmarks in 16 African languages and English. On
AfriMMLU, token-probability confidence ranks answers far worse in African languages than in
English, with an area under the ROC curve (AUROC) of $0.60$ vs.\ $0.82$.
Restricting English to its hardest questions until accuracy matches almost removes this gap,
because English confidence also ranks those questions poorly. At equal English difficulty most of
the gap remains, but African accuracy is lower at almost every level, so this comparison cannot
separate weaker confidence from questions that are harder for the model. On AfriMMLU, verbalised
confidence, stated in words, is worse calibrated than token probability in African languages and
not in English. A model held to its English error rate can answer only $7.6\%$ ($95\%$ interval
$5.1$--$10.4\%$) of African AfriMMLU questions, and recalibration that keeps the ranking does not
raise it. We recommend that such comparisons report a difficulty-adjusted gap alongside
the raw one and name the English questions and difficulty score used.
\end{abstract}

\begin{figure}[t]\centering
\includegraphics[width=\columnwidth]{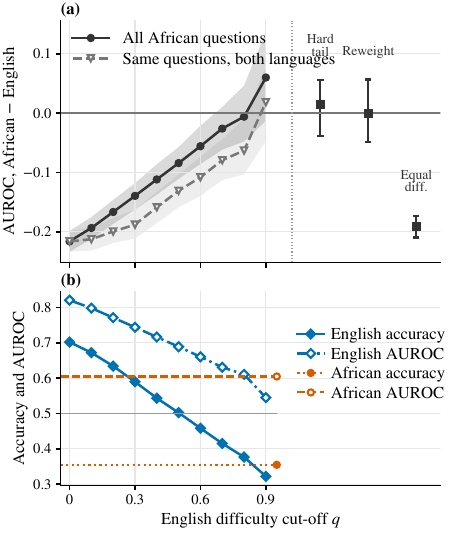}
\caption{How the African--English AUROC difference on AfriMMLU depends on the English
questions used (token probability, seven models). \textbf{(a)} The difference when English is
restricted to the questions at or above difficulty quantile $q$, for all African questions (solid)
and for the same questions in both languages (dashed), with $95\%$ bootstrap bands. Difficulty is
the error rate of $5{,}226$ leaderboard models \citep{kipnisMetabenchSparseBenchmark2025}. The
points on the right are three adjusted comparisons (\S\ref{sec:prelim}): English restricted to its
hardest questions until its accuracy matches the African accuracy (hard tail), all English
questions reweighted towards hard ones until accuracy matches, and the comparison at equal English
difficulty. \textbf{(b)} Accuracy and AUROC of the English questions kept, with the African values
for reference.}
\label{fig:overview}
\end{figure}

\section{Introduction}
\label{sec:intro}

Large language models (LLMs) are increasingly proposed for education, health information,
legal and government services in African languages, where confidently wrong answers can carry
real costs. Africa's more than two thousand languages represent about a third of the world's
\citep{adebaraSERENGETIMassivelyMultilingual2023}, and most are under-resourced for natural
language processing (NLP) \citep{adebaraAfrocentricNLP2022}. Community efforts around Masakhane
\citep{nekotoParticipatoryResearchLowresourced2020} have produced benchmarks including
IrokoBench \citep{adelaniIrokoBenchNewBenchmark2024} and Uhura
\citep{bayesUhuraBenchmarkEvaluating2024}, which document large accuracy gaps between English
and African languages. Accuracy alone does not reveal when to trust an answer. A deployed model
can instead \emph{defer}, passing the questions it is least sure of to a human or another
system. Deferral needs confidence that discriminates, ranking correct answers above incorrect
ones \citep{geifmanSelectiveClassificationDeep2017}, and a confidence threshold is easier to set
when confidence is also calibrated, so that a confidence of $0.8$ means that about $80\%$ of such
answers are correct \citep{guoCalibrationModernNeural2017}.

Confidence is hard to compare across languages. Studies of multilingual calibration mostly
compare the same translated questions in every language, at whatever accuracy each language
reaches
\citep{ahujaCalibrationMassivelyMultilingual2022,yangCalibrationMultilingualQuestion2023,xueMlingConf2024,zhouFinalLayerIntermediate2025}.
Accuracy is lower in less-resourced languages, where usable web text is scarce
\citep{kreutzerQualityGlance2022}, and confidence may rank answers less well on harder questions
in any language. A raw difference in how well confidence ranks answers therefore mixes two
explanations. The model's confidence may be weaker in the African language, or the questions may
simply be harder for the model in that language.

We compare how well confidence ranks answers in 16 African languages and in English while
changing which English questions the African languages are compared with. We use all English
questions, then only the harder ones, then only as many hard questions as are needed to match the
African accuracy. Finally, we compare the languages at the same English difficulty, measured by
how often other models get each question wrong (Fig.~\ref{fig:overview}). We make these
comparisons for token probability (the probability the model assigns to its answer) and
verbalised confidence (a confidence it states in words), and we evaluate a third signal,
self-consistency (the agreement among several sampled answers). We then measure what the gap costs a deployer, as the share of
African-language questions a model can answer, most confident first, while its error rate stays
within its English error rate.

\begin{figure*}[t]\centering
\includegraphics[width=\textwidth]{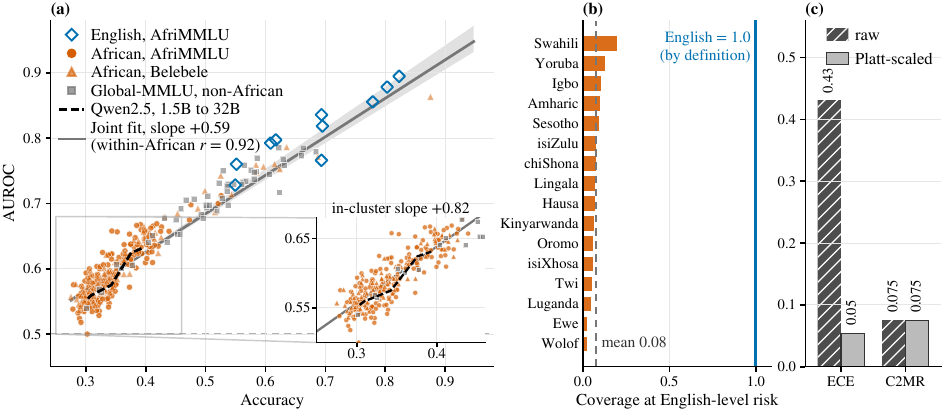}
\caption{Accuracy and AUROC, coverage at English-level risk, and recalibration (token
probability). \textbf{(a)} Each point is one model in one language. English results (open
diamonds) and African results (filled circles) lie close to one curve, fitted over all $287$
model--language pairs (slope $+0.59$), with a $95\%$ bootstrap band over those pairs. The main cloud is AfriMMLU
(ten models), with the Qwen2.5 models from 1.5B to 32B connected. Triangles are Belebele
\citep{belebele2024} (eight models), and squares are six further non-African languages from
Global-MMLU \citep{singhGlobalMMLUUnderstanding2024}, which are not part of the fit. The inset
shows a separate fit to the dense cluster. \textbf{(b)} The share of questions a model can answer
in each African language while its error rate stays within its English error rate
(coverage-to-match-risk, seven models). \textbf{(c)} Recalibrating each language with Platt
scaling lowers African expected calibration error (ECE) and leaves this share unchanged.}
\label{fig:auroc-acc}
\end{figure*}

We find, first, that the African--English gap in ranking depends on which English questions are
used for the comparison. On AfriMMLU, matching English to the African accuracy brings the gap
close to zero, because the English questions that remain are ones on which English confidence
also ranks poorly. At equal English difficulty most of the gap remains, although the African
questions are still answered less accurately at almost every level of difficulty. Second,
verbalised confidence is worse calibrated than token probability in the African languages on
AfriMMLU and not in English. Third, a model held to its English error rate can answer only a small
share of African questions, well below what a perfect ranking of the same answers would allow.
Recalibration that keeps the ranking of answers does not raise that share
(Fig.~\ref{fig:auroc-acc}).

A raw gap mixes the two explanations, and matching accuracy can hide the gap. A gap that
remains at equal English difficulty still cannot be attributed to either explanation while
accuracy differs. We therefore propose that such comparisons report a difficulty-adjusted gap next
to the raw one, and say which English questions and which difficulty score they rely on. We
release the evaluation code, each model's confidence and correctness on every question, and the results of every analysis so that others
can repeat the analyses.\footnote{Code and data are available at \url{https://github.com/sgrimbly/africal-calibration}.}

\section{Related Work}
\label{sec:related}

\paragraph{Calibration and selective prediction.}
The expected calibration error (ECE) \citep{pakdamannaeiniObtainingWellCalibrated2015} summarises
miscalibration, the Brier score \citep{brierVERIFICATIONFORECASTSEXPRESSED1950} measures overall
probabilistic error, and
temperature scaling is a simple and effective post-hoc fix \citep{guoCalibrationModernNeural2017}.
In selective prediction a model abstains on some questions, trading the share it answers
(coverage) for the error rate on those it answers (selective risk)
\citep{geifmanSelectiveClassificationDeep2017}. Coverage at a target risk is a standard
operating point on the risk--coverage curve
\citep{geifmanSelectiveClassificationDeep2017,kamathSelectiveQuestionAnswering2020,whiteheadReliableVQA2022}.
\emph{Equalized coverage} in conformal prediction \citep{romanoMaliceTowardNone2020} fixes a
coverage guarantee across groups, and selective classification can magnify accuracy
disparities between groups \citep{jonesSelectiveClassificationDisparities2021}. Concurrent work
finds that one confidence threshold accepts English requests more often than requests in other
languages, and that English keeps more coverage when each language has its own threshold at a
common selective risk \citep{yaoOneConfidenceThreshold2026}. Methods for deciding when to abstain
include multilingual self-feedback \citep{fengTeachingLLMsAbstain2024}, cross-lingual answer
consensus \citep{mkaMultilingualAbstain2025} and causal selection of feedback
\citep{causalAbstain2025}. Methods that abstain by confidence need a signal that ranks answers
well, which raises the question of how confidence is obtained.

\paragraph{Confidence in language models.}
Post-hoc correction improves the calibration of generative question answering (QA)
\citep{jiangHowCanWe2020}, and models can learn to predict whether their own answers are
correct \citep{kadavathLanguageModelsMostly2022}. Confidence can be read from token
probabilities, stated in words as verbalised confidence
\citep{linTeachingModelsExpress2022,xiongCanLLMsExpress2023},
or derived from agreement across sampled answers \citep{wangSelfConsistencyImprovesChain2022}
\citep[see][for a survey]{gengSurveyConfidenceEstimation2023}. For models tuned with
reinforcement learning from human feedback (RLHF), verbalised confidence is often better calibrated
than the model's conditional probabilities \citep{tianJustAskCalibration2023}. In
English, what a model says about its confidence can disagree with what its internal signals
carry, and recent methods aim to reduce this mismatch
\citep{liuMetaFaith2025,miaoClosingConfidenceFaithfulness2026}. Other English-only work
separates how well confidence ranks answers from accuracy \citep{cacioliMetacognitiveEfficiency2026}
and tests whether confidence stays stable under paraphrased prompts and equivalent answers
yet changes when an answer's meaning does \citep{xiaCalibrationNotEnough2026}.
ConfidenceBench evaluates stated probabilities from 15 frontier models on English questions
\citep{ffrenchConstantConfidenceBench2026}, and \citet{steyversCalibrationGap2024} study how
far people's trust in a model's answers differs from what the model knows. Across languages
the evidence is thinner. \citet{rathiHumansOverrely2025} study the markers of certainty and uncertainty that LLMs
attach to their answers in five high-resource languages, and find that expressions of high
certainty often accompany wrong answers. They also show that people rely heavily on such markers.
\citet{fengTeachingLLMsAbstain2024} find that verbalised confidence can collapse to near-constant
values in low-resource languages. On the low-resource languages of multilingual MMLU (Massive Multitask Language
Understanding) \citep{hendrycksMeasuringMassiveMultitask2020}, Aya-13B also abstains less
accurately with verbalised confidence than with token probability. For token probabilities,
\citet{huangMultilingualCalibrationInstructionTuning2026} find
that instruction tuning raises confidence in many languages, including low-resource ones absent
from the tuning data, with little or no gain in accuracy. We evaluate one signal of each kind and
compare them across languages.

\paragraph{Multilingual and African NLP.}
Calibration is worse in lower-resource languages
\citep{ahujaCalibrationMassivelyMultilingual2022,yangCalibrationMultilingualQuestion2023}.
These studies compare languages at whatever accuracy each reaches, as do comparisons of
confidence for GPT-3.5 closed-book QA across high- and low-resource settings
\citep{krauseConfidentlyWrong2023} and across the five main high- and mid-resource languages in
MlingConf \citep{xueMlingConf2024}. Concurrent work compares the area under the receiver
operating characteristic curve (AUROC) across languages on parallel items and treats them as
equally difficult, although it finds that the translation source itself changes accuracy
\citep{duLostTranslationNot2026}. Calibration metrics that control for accuracy have been used to
compare models rather than languages \citep{yangCalibrationRankingsReverse2026}, and a
22-language study of confidence from reasoning holds accuracy fixed only within a language
\citep{alfaranoEstimatingUncertaintyReasoning2026}. Reports on African benchmarks and
African-focused models compare accuracy between English and African languages
\citep{adelaniIrokoBenchNewBenchmark2024,ojoAfroBenchHowGood2023} and between better- and
less-resourced African languages \citep{adebaraWhereAreWe2025}, but do not report confidence
quality. Many of the languages we study already appear in prior work, such
as Amharic in \citet{krauseConfidentlyWrong2023}, Swahili in
\citet{ahujaCalibrationMassivelyMultilingual2022} and \citet{yangCalibrationMultilingualQuestion2023},
and 14 of the 16 in the Belebele evaluation of \citet{zhouFinalLayerIntermediate2025}. That
study, the closest to ours, reports ECE, Brier score and AUROC across more than 100 languages
and relates confidence quality to resource level, without adjusting for the lower non-English
accuracy. A hackathon report finds higher ECE in African languages than in English on AfriMMLU
and Uhura-TruthfulQA for three of four open-weight models, and that per-language temperature scaling
removes much of the calibration error \citep{okunowoConfidentlyWrong2026}. We know of no cross-lingual comparison of confidence ranking
that adjusts for item difficulty beyond translating the same questions, or that examines how the
result depends on which English questions are used.

\section{Measures and Reference Sets}
\label{sec:prelim}

We study multiple-choice questions. For an item with correct answer $y$, a model
returns a predicted label $\hat{y}$ and a scalar confidence
$\hat{p}\in[0,1]$. The correctness indicator is $c = \mathbf{1}[\hat{y}=y]$,
which we treat as a Bernoulli outcome whose probability the confidence should
predict. A confidence signal has three distinct properties, and each can fail without the others.
\emph{Calibration} asks whether confidence matches empirical accuracy at that level
(ECE). \emph{Discrimination} asks whether confidence ranks correct predictions
above incorrect ones, irrespective of the accuracy level (AUROC, where chance is $0.5$).
\emph{Selective prediction} asks how much abstaining on low-confidence items lowers the error rate.
A model answering at $30\%$ accuracy while reporting $30\%$ confidence on every item, for
example, has near-zero ECE but chance AUROC.

We estimate miscalibration by
$\mathrm{ECE} = \sum_{m}\frac{|B_m|}{n}\,|\operatorname{acc}(B_m)-\operatorname{conf}(B_m)|$
over $15$ equal-mass bins $\{B_m\}$ (adaptive ECE). Empirical ECE is binning-sensitive, and
equal-width bins are biased when confidence is concentrated or samples are small
\citep{vaicenaviciusEvaluatingModelCalibration2019,mindererRevisitingCalibrationModern2021,nixonMeasuringCalibrationDeep2019}.

Answering only the most confident $\kappa$ fraction yields coverage $\kappa$ and selective risk
$r(\kappa)$, the error rate over the answered set, and the risk--coverage curve traces $r$ over
$\kappa$ \citep{geifmanSelectiveClassificationDeep2017}. The appendix also reports the
excess area under the risk--coverage curve (\emph{excess-AURC}), the area between this curve and
the best curve attainable at the observed accuracy \citep{geifmanBiasReducedUncertainty2019}. As a measure of discrimination that does not
depend on the share of correct answers, we use the AUROC of confidence for predicting
correctness. It is the probability that a randomly chosen correct prediction is more confident
than a randomly chosen incorrect one, with ties counted as one half. Because AUROC and excess-AURC depend only on the confidence
ranking, any strictly increasing recalibration map, however rich \citep{kullBetaCalibration2017},
leaves both unchanged.

To measure the cost of the gap for deployment, let $r^{\mathrm{en}}_{\text{full}}$ be a model's
selective risk at full coverage in English. For a target language $\ell$ with risk--coverage
curve $r_\ell(\kappa)$, we define \emph{coverage-to-match-risk} (C2MR) as
\begin{equation}
\mathrm{C2MR}(\ell) = \max\{\kappa : r_\ell(\kappa)\le r^{\mathrm{en}}_{\text{full}}\},
\end{equation}
the largest share of questions in $\ell$ the model can answer, abstaining on the least
confident, while its error rate stays within its English error rate. A value of $1$ means $\ell$
is as safe to answer as English, and $0.4$ means abstaining on the least confident $60\%$. The
maximum runs over distinct confidence thresholds, and $\mathrm{C2MR}=0$ if no threshold keeps the
error rate within the English one. Equalized coverage targets the same prediction-interval coverage in every
group \citep{romanoMaliceTowardNone2020}, whereas C2MR fixes the risk at the English level and
reads off the coverage each language attains. C2MR is a retrospective diagnostic. Its threshold
is chosen on the same outcomes it is scored on, and it reports achievable coverage in hindsight.
A deployable version would choose the threshold on held-out data and report risk and coverage on
new questions, as in selective classification with a guaranteed risk
\citep{geifmanSelectiveClassificationDeep2017}.

The raw African--English difference is computed on translations of the same questions,
answered at very different accuracies. We therefore vary the English questions used for the
comparison, which we call the \emph{English reference set}, and report how the difference
changes. For a metric $M$ the difference is
$\Delta_M = M(F^{\mathrm{Afr}}) - M(F^{\mathrm{Eng}})$, where $F^{\mathrm{Afr}}$ holds a model's
answers in the African languages and $F^{\mathrm{Eng}}$ its answers on the English reference
set. The methods below all use a difficulty score for each English question. Because AfriMMLU's
English questions are MMLU test questions, we score each by its error rate across up to $5{,}226$
open models from the Open LLM Leaderboard recorded in the metabench response data
\citep{kipnisMetabenchSparseBenchmark2025}. None of the models we evaluate are among them. We
call this the leaderboard score. A second score comes from the errors of the other models we
evaluate, leaving out the model being scored. AfriXNLI has no leaderboard record, so only the
second score is available there.

We use the scores in three ways. First, we keep the English questions at or above difficulty
quantile $q$, for $q=0,0.1,\dots,0.9$, and compare them with all African questions and,
separately, with the same questions in the African languages. Two methods then match English
accuracy to the African accuracy. For the \emph{hard tail}, we add English questions from hardest
to easiest until English accuracy reaches the African level. For the \emph{difficulty
reweighting}, we keep every English question but give harder questions more weight, until the
weighted English accuracy matches the African accuracy. The weights depend only on the difficulty
score, and each subject keeps its share of the questions used. Finally, for the \emph{equal-difficulty
comparison}, we compute the English AUROC and one AUROC over all African-language answers within
equal-size deciles of English difficulty, and average the deciles with equal weight, so that both
languages are compared on the same difficulty distribution. Deciles in which either AUROC is
undefined are left out. This is a stratified, covariate-adjusted
AUROC \citep[cf.][]{janesAdjustingCovariateEffects2009}. Janes and Pepe weight the strata by how
the correct answers are distributed over them. As variants, we weight each decile by its number
of correct--incorrect pairs in English or in the African languages. We report these methods as
sensitivity analyses of the reference set.

As a control, we discard correct English answers at random until English accuracy matches the
African accuracy. This leaves the confidence distributions of correct and of incorrect answers
unchanged in expectation. Because AUROC depends only on those class-conditional distributions
\citep{fawcettIntroductionROC2006}, the subsample reproduces the full-English difference in
expectation. Any change under the methods above therefore comes from which questions are kept.
Intervals for these comparisons resample question IDs within subject and use the same draw
across models. Every selection, cut-off and weighting is re-fitted inside each replicate, with
the difficulty scores and deciles held fixed.

\section{Experimental Setup}
\label{sec:setup}

We evaluate 16 African languages and English, grouped by resource tier following
\citet{joshiStateFateLinguistic2020} (Table~\ref{tab:langs}). The languages are split into more
tokens than English for the same text \citep{petrovTokenizersUnfairness2023,ahiaAllLanguagesCost2023}
and have little usable web text \citep{kreutzerQualityGlance2022}, which makes them a demanding
test of confidence. The two primary tasks come from IrokoBench
\citep{adelaniIrokoBenchNewBenchmark2024}. AfriMMLU has $500$ four-option questions per language, professionally translated from
MMLU test questions in five subjects, and AfriXNLI is three-option natural language inference
with $600$ items per language. Uhura ARC-Easy \citep{bayesUhuraBenchmarkEvaluating2024} serves as
a cross-benchmark check, and Belebele \citep{belebele2024} and Global-MMLU
\citep{singhGlobalMMLUUnderstanding2024} add results at higher accuracy and further languages
(Table~\ref{tab:repro-tasks}). AfriMMLU, AfriXNLI and Uhura are translated from English by humans.
AfriMMLU and AfriXNLI are aligned question by question, which lets us compare the same question
across languages. The Uhura test sets cover different subsets of the English questions. The Ewe, Lingala
and Wolof AfriMMLU sets were translated through French \citep{adelaniIrokoBenchNewBenchmark2024}.
Each question is asked once for every cyclic shift of its answer options, four times on AfriMMLU
and three on AfriXNLI, so that every option appears in every position and position bias averages
out.

The models we evaluate (Table~\ref{tab:roster}) include models used for African languages,
three African-tuned models, and the Qwen2.5 family at five sizes. The three
African-tuned models (InkubaLM, Lugha-Llama, AfroLlama-V1) score at or below multiple-choice
chance even in English, so their confidence cannot be assessed, and we leave them
out of the analyses. The other ten models, from the Qwen, Llama, Gemma and Aya families, score
above chance on AfriMMLU and reach a sensible English accuracy on MMLU. They include two 32B
models quantised to 4 bits with activation-aware weight quantisation (AWQ)
\citep{linAWQActivationawareWeight2024}, added to cover high accuracy. Most analyses use the
seven of these ten for which we collected both token-probability and verbalised confidence:
Qwen2.5-7B and 32B, Qwen3-8B and 32B, Llama-3.1-8B, Gemma-2-9B and Aya-Expanse-8B.
Self-consistency was run on the eight models below 32B, and the Belebele points of
Fig.~\ref{fig:auroc-acc}a use the same eight.

We elicit confidence in three ways. For token probability, we normalise the first-token
log-probabilities of the option letters, which needs log-probability access
(Table~\ref{tab:signal-taxonomy}). Verbalised confidence comes from a separate prompt that asks for
an answer and a confidence, so the two prompts can give different answers. The prompt follows the
zero-shot protocol of \citet{tianJustAskCalibration2023}, with percentages as in
\citet{linTeachingModelsExpress2022}, and we parse both the answer and the confidence. For self-consistency, we sample $n=10$
single-letter answers at temperature $0.7$ and use the share of parsed answers that agree with the
most common one, adapting the consistency measures of \citet{wangSelfConsistencyImprovesChain2022}
and \citet{xiongCanLLMsExpress2023} to answers sampled without reasoning chains. All prompts are in English,
and prompting in the target language is left to future work. In a \emph{translate-test}
condition we machine-translate the African questions into English with the NLLB-200 translation
model \citep{nllbteamNoLanguageLeft2022}, to test whether the gap comes from reading the question
in an African language.

Platt scaling \citep{plattProbabilisticOutputs1999}, a logistic map of the logit of confidence
with a positive slope, fitted for each language on a held-out split, is the recalibration
baseline. Intervals are percentile bootstraps ($\alpha=0.05$) that resample question IDs, because
each question appears in several option orders and languages and resampling individual rows
would understate the variance. Intervals for a single model and language use $B=1000$, and
pooled differences use $B=2000$--$10000$.
Inference runs on an offline vLLM backend \citep{kwonEfficientMemoryManagement2023}.

\section{Results}
\label{sec:results}

\subsection{Confidence ranking in African languages and English}
\label{sec:c1}

\begin{table*}[t]
\centering\footnotesize\setlength{\tabcolsep}{3pt}
\begin{tabular}{@{}>{\raggedright\arraybackslash}p{4.5cm}llll@{}}
\toprule
& \multicolumn{3}{c}{Token probability} & Verbalised \\
\cmidrule(lr){2-4}\cmidrule(l){5-5}
& AfriMMLU & AfriMMLU & AfriXNLI & AfriMMLU \\
Difficulty from & leaderboard models & other models & other models & leaderboard models \\
\midrule
AUROC, African vs.\ English & $.60$ vs.\ $.82$ & $.60$ vs.\ $.82$ & $.58$ vs.\ $.78$ & $.54$ vs.\ $.71$ \\
African minus English, all questions & $-.216$ {\scriptsize$[-.233,-.198]$} & $-.216$ {\scriptsize$[-.233,-.198]$} & $-.206$ {\scriptsize$[-.231,-.184]$} & $-.173$ {\scriptsize$[-.196,-.149]$} \\
\midrule
\multicolumn{5}{@{}l}{\textit{English matched to African accuracy}} \\
Hardest English questions only (hard tail) & $+.015$ {\scriptsize$[-.039,+.056]$} & $+.024$ {\scriptsize$[.000,+.056]$} & $+.097$ {\scriptsize$[+.042,+.152]$} & $-.033$ {\scriptsize$[-.095,+.023]$} \\
\hspace{0.8em}the same questions in both languages & $-.046$ {\scriptsize$[-.101,-.001]$} & $-.062$ {\scriptsize$[-.091,-.029]$} & $+.059$ {\scriptsize$[+.003,+.119]$} & $-.067$ {\scriptsize$[-.123,-.004]$} \\
All questions, reweighted towards hard ones & $-.001$ {\scriptsize$[-.048,+.057]$} & $-.048$ {\scriptsize$[-.079,-.003]$} & $+.061$ {\scriptsize$[+.009,+.118]$} & $-.021$ {\scriptsize$[-.112,+.019]$} \\
\midrule
\multicolumn{5}{@{}l}{\textit{Equal English difficulty, with difficulty deciles weighted}} \\
\hspace{0.8em}equally & $-.191$ {\scriptsize$[-.209,-.172]$} & $-.157$ {\scriptsize$[-.173,-.137]$} & $-.216$ {\scriptsize$[-.231,-.195]$} & $-.118$ {\scriptsize$[-.139,-.087]$} \\
\hspace{0.8em}by English correct--incorrect pairs & $-.158$ {\scriptsize$[-.183,-.134]$} & $-.116$ {\scriptsize$[-.140,-.094]$} & $-.125$ {\scriptsize$[-.163,-.089]$} & $-.104$ {\scriptsize$[-.130,-.076]$} \\
\hspace{0.8em}by African correct--incorrect pairs & $-.196$ {\scriptsize$[-.217,-.171]$} & $-.168$ {\scriptsize$[-.189,-.143]$} & $-.217$ {\scriptsize$[-.246,-.185]$} & $-.124$ {\scriptsize$[-.147,-.088]$} \\
\bottomrule
\end{tabular}
\caption{African minus English AUROC for the seven models when the English questions
are restricted or reweighted to match African accuracy, or compared at equal English difficulty.
Difficulty comes from the errors of $5{,}226$ leaderboard models (AfriMMLU only) or of the other
evaluated models. Brackets are $95\%$ bootstrap intervals that resample questions ($B=2000$).
The reweighting solver finds a tilt that matches African accuracy in $1{,}317$ and $1{,}935$ of the $2{,}000$ replicates
on AfriMMLU with token probability (leaderboard and other models), $1{,}767$ on AfriXNLI and
$927$ with verbalised confidence, and its intervals use those replicates. The verbalised column
keeps a question when every model has at least one parsed verbalised reply in English and one in
the African languages. This leaves $389$ questions, $9$ of them from elementary mathematics,
compared with $479$ and $79$ for token probability. Over all ten models
with token probability, the gap on all questions stays between $-0.196$ and $-0.251$ when any one
subject is left out, when the questions MMLU-Redux \citep{gemaAreWeDone2025} flags as
erroneous are removed, or when the languages are pooled in any of three ways. Discarding
correct English answers at random until accuracy matches leaves it at $-0.223$. Deciles in
which a model answers every English question correctly have no English AUROC and are left out.
Table~\ref{tab:refset} gives the ten-model values.}
\label{tab:c1}
\end{table*}

Confidence ranks answers far better in English than in the African languages. On AfriMMLU,
across the seven models, AUROC is $0.82$ in English and $0.60$ in the African languages
(Table~\ref{tab:c1}), and African accuracy is half the English accuracy. AfriXNLI shows the same
gap, and English is also better calibrated on both tasks. Confidence in the African languages
still carries some information, and on AfriMMLU it ranks answers above chance for almost every
model and language. The smaller share of correct answers does not produce the gap by itself,
since discarding correct English answers at random until the accuracies match leaves it
unchanged. Uhura ARC-Easy shows a larger gap in the same direction, but its English accuracy is
so high that too few English errors remain for the comparisons that follow.

The gap narrows as the English questions are restricted to harder ones (Fig.~\ref{fig:overview}a),
because English confidence itself ranks hard questions poorly (Fig.~\ref{fig:overview}b). On the
hardest tenth of English questions, English AUROC falls to $0.54$, below the $0.60$ of the
African questions. With the leaderboard score no restriction makes the gap reliably positive, but
with difficulty from the other evaluated models the hardest tenth does
(Appendix~\ref{app:refset}).

Matching English to African accuracy brings the gap close to zero. This holds for both the hard
tail and the reweighting, with a sign that depends on the method and on the difficulty score
(Table~\ref{tab:c1}). When the African languages are restricted to the same hard-tail questions,
a small gap in favour of English remains. With difficulty from the other evaluated models, the hard-tail
difference slightly favours the African languages. On the ten models, seven of which are Qwen
models, this advantage disappears when the Qwen models are left out of the score. Calibration changes sign
under the same restriction. On all questions the African languages are worse calibrated than
English (ECE higher by $0.19$). Compared with English on the hard tail they are better
calibrated (ECE lower by $0.09$), because the models answer many of the hardest English questions
confidently and incorrectly.

Comparing the two languages at equal English difficulty keeps most of the gap. The gap is
negative for each of the seven models. Removing the shared questions that shrink it most, one at
a time, takes about a third of them to bring it to zero.
The two difficulty scores may err in opposite directions. The leaderboard score predicts each
evaluated model's errors less well and may under-correct for difficulty. The other evaluated
models share many of that model's errors, and their score may over-correct. On AfriMMLU with token
probability, between $54\%$ and $91\%$ of the gap remains across both scores and the three ways
of weighting the deciles (Table~\ref{tab:c1}).

Equal English difficulty does not equalise accuracy. In almost every difficulty decile the
African questions are still answered less accurately, and across models and deciles the
difference in AUROC tracks the difference in accuracy ($r=0.86$), with a fitted difference near
zero where the accuracies are equal. Where the two accuracies are close, both are near chance and
confidence ranks poorly in both languages.

AfriXNLI and verbalised confidence show a similar pattern (Table~\ref{tab:c1}). Matching accuracy
reverses the gap on AfriXNLI and brings it close to zero for verbalised confidence, and most or
all of it remains at equal English difficulty.
Self-consistency behaves like token probability, with African calibration better than English
calibration on the hard tail for each of the eight models below 32B (Fig.~\ref{fig:forest}),
and four times as many samples raise African AUROC by at most $0.024$ for three models. Results for the 32B models alone and for model--language
pairs with accuracy of at least $0.50$ are also in the appendix.

\subsection{Model size and translation}
\label{sec:scale}

In both Qwen series the gap is wider at 32B than at 7--8B (Table~\ref{tab:main}), because size
raises English accuracy far more than African accuracy. Across models and languages, AUROC rises
with accuracy along one curve that English and African results share
(Fig.~\ref{fig:auroc-acc}a). From Qwen2.5-1.5B to 32B, English accuracy rises from $0.55$ to
$0.82$ and African accuracy only from $0.30$ to $0.39$, so English moves much further up the
curve. The shared curve does not show that confidence ranks answers equally well in English and
the African languages at equal difficulty. Its points come from different models and questions,
and the equal-difficulty comparison of \S\ref{sec:c1} still finds a gap. A similar curve appears
on Belebele \citep{belebele2024}, a reading-comprehension benchmark built from a different
English source.

Translating the African questions into English closes part of the gap.
\label{sec:transient}
For the eight models below 32B, translation raises accuracy by $0.09$ and AUROC by $0.06$
compared with the untranslated questions. At equal English difficulty, on the $446$ questions
that every translated and untranslated run answered, the gap is $-0.14$ for the translated
questions and $-0.18$ for the untranslated ones, so translation closes about a quarter of it. The
gap is also smaller where less of the question has to be read in the African language. On
elementary-mathematics questions, whose answer options are mostly numbers, the raw
gap is about half as large.

\subsection{Verbalised confidence}
\label{sec:c2}

\begin{figure}[t]\centering
\includegraphics[width=\columnwidth]{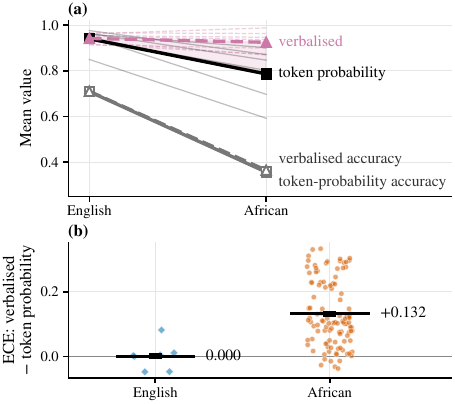}
\caption{Confidence under the verbalised and token-probability prompts.
\textbf{(a)} Mean confidence and accuracy under each prompt, for English and the African languages. Faint lines are individual
models and bold lines are group means. \textbf{(b)} Verbalised minus token-probability ECE, each evaluated
on its own answers, on AfriMMLU questions where both responses parse. Each point is one model in
one language. The model can answer differently under the two prompts.}
\label{fig:sayknow}
\end{figure}

On AfriMMLU, verbalised confidence is worse calibrated than token probability in the African
languages and about as well calibrated in English (Fig.~\ref{fig:sayknow}). Because the two
prompts can choose different answers, we evaluate each prompt's confidence on its own answers. On
questions where both prompts give a parseable answer, verbalised ECE in the African languages is
$0.561$ and token-probability ECE $0.429$, a difference of $0.132$ $[0.126,\allowbreak 0.137]$. In English
the two are essentially equal. The extra error is positive for six of the seven models and remains on
questions where both prompts choose the same answer. On AfriXNLI it is small.

The two prompts choose the same option on $73\%$ of African AfriMMLU questions, and verbalised
answers fail to parse more often. Counting failed parses as wrong, the verbalised prompt is less
accurate in the African languages than the token-probability prompt, while on questions where
both parse it is slightly more accurate (Table~\ref{tab:verb-prompts}). Verbalised confidence also ranks answers less
well than token probability in both languages (Table~\ref{tab:c1}).

\subsection{Coverage at English-level risk}
\label{sec:c3}

At English-level risk, a model can answer only a small share of African questions. Across
the seven models and the African languages, coverage-to-match-risk is $0.076$
$[0.051,\allowbreak 0.104]$ (Fig.~\ref{fig:auroc-acc}b), compared with $1.0$ in English by
definition. With the same answers, a perfect ranking would reach $0.508$, so accuracy and ranking
each account for about half of the shortfall. The operating point lies in the extreme tail of the
confidence distribution, where small differences in ranking decide coverage. Per language,
coverage follows accuracy more closely than resource tier (Fig.~\ref{fig:c2mr}).

Recalibration repairs calibration but leaves coverage unchanged. Platt scaling each language
within each model, fitted and evaluated on disjoint halves, lowers African ECE from $0.43$ to about
$0.05$ (Fig.~\ref{fig:auroc-acc}c). Because the fitted map is strictly increasing, it cannot
reorder confidence, and coverage stays at $0.075$ on the evaluation halves
(Table~\ref{tab:recal}). The same holds for verbalised confidence, whose ECE falls to $0.04$ while
its AUROC stays near chance. Self-consistency reaches zero coverage at English-level risk for
the three models tested with both $n=10$ and $n=40$ samples.

\section{Discussion and Conclusion}
\label{sec:conclusion}
We asked how the African--English gap in confidence ranking depends on the English questions
used for the comparison. On all questions the gap is large. Matching English to the African
accuracy brings it close to zero on AfriMMLU, and comparing the languages at equal English
difficulty keeps most of it.

The two kinds of adjustment answer different questions. Matching English to African accuracy
compares African confidence with English confidence on the hardest English questions, where
English confidence ranks answers about as poorly as African confidence does. Comparing within
deciles of equal English difficulty removes what confidence gains from telling easy questions
from hard ones. The African questions are still answered less accurately in almost every
decile, and the difference in AUROC follows the difference in accuracy. We therefore cannot say whether the
remaining gap comes from weaker confidence or from questions that are harder for the model in the
African language. Either way, an accuracy-matched comparison reported alone can hide the gap, and
a raw comparison reported alone does not show how much of it depends on difficulty.

The extra calibration error of verbalised confidence on AfriMMLU could come from how the
model reports confidence or from which answers the verbalised prompt leads it to choose. Fixing
the answer first and then asking for confidence would separate the two, and we leave a full
test to future work. \citet{tianJustAskCalibration2023} find verbalised confidence better calibrated than the
model's conditional probabilities for closed RLHF-tuned models on English short-answer QA, with
mixed results for the open-weight Llama-2-70B-Chat. We do not observe this advantage. In open-ended QA and
arithmetic with one model and six or seven non-African languages,
\citet{zhouFinalLayerIntermediate2025} find perplexity-based, margin-based and verbalised
confidence all worse calibrated outside English, but find the excess error of verbalised over
perplexity- and margin-based confidence no larger there.

For a deployer, low African accuracy caps coverage at English-level risk at about half of the
questions, and the observed ranking lowers it to $0.076$. Because coverage depends on the
ranking, calibration results on their own do not show whether confidence can be relied on for
deferral in these languages. Comparisons of confidence across languages should give the raw gap
together with a gap adjusted for difficulty, and name the English questions and the difficulty
score behind the adjustment.

\section*{Limitations}
\label{sec:limitations}
We measure confidence on multiple-choice questions, so the probability of an option letter is a
proxy for confidence in a full answer and may not transfer to open generation. Verbalised
confidence is also observed only when the response can be parsed, and our audit of the generated
replies shows that this selection reverses which prompt is more accurate in the African
languages on AfriMMLU. We have not
rerun the matched verbalised analysis with imputed confidence for failed responses
(Appendix~\ref{app:parse}).

Every comparison at matched accuracy or equal difficulty rests on a difficulty score defined in
English, and ``equal difficulty'' in this paper means equal English difficulty. That score
transfers poorly to the African languages. Across questions, English and African accuracy
correlate at only $\rho=0.35$ on AfriMMLU and $0.15$ on AfriXNLI. The leaderboard score comes
from older leaderboard models answering a different prompt format. The score from the other
evaluated models inherits the errors of the Qwen models. Across these choices and the decile
weightings, the equal-difficulty difference on AfriMMLU ranges from $-0.116$ to $-0.196$
(Table~\ref{tab:c1}), and AfriXNLI has only the second score. The hard tail and the reweighting
also use the evaluated model's English outcomes to decide when accuracy is matched.

Every benchmark here was translated from English, and the claims apply to translated benchmarks.
Tasks written natively in African languages may behave differently. Seven of the ten models are
Qwen models and the two 32B models are quantised, so ``four families'' overstates their
diversity. Because we ran no deployment study, the operational claims describe error-rate targets
computed on benchmark data.

\section*{Ethics Statement}
This study collects no new data from people. All benchmarks are public releases used under their
licences, and we cite the translators and annotators whose work they contain. Confidence that
ranks answers poorly exposes speakers of these languages to confidently wrong answers
\citep[cf.][]{frickerEpistemicInjustice2007}, so it should be measured before these models are
used to decide when to defer.

\bibliography{references}

\clearpage
\appendix
\renewcommand{\topfraction}{0.9}\renewcommand{\bottomfraction}{0.8}
\renewcommand{\textfraction}{0.07}\renewcommand{\floatpagefraction}{0.85}
\renewcommand{\dbltopfraction}{0.9}\renewcommand{\dblfloatpagefraction}{0.85}
\setcounter{topnumber}{4}\setcounter{dbltopnumber}{3}\setcounter{totalnumber}{6}
\newcommand{\estci}[3]{$#1\,[#2,#3]$}
\FloatBarrier
\section*{Appendix}
\noindent The appendix has four parts. Appendix~\ref{app:c1} gives results for each model and each
language, a test on Belebele at higher accuracy, and coverage at English-level risk with
recalibration. Appendix~\ref{app:refset} gives the details of the English reference sets and the
difficulty scores, and results above several accuracy thresholds. Appendix~\ref{app:verb}
audits the comparison of verbalised and token-probability confidence. Appendix~\ref{app:repro} gives
the setup and reproducibility details. We call one model evaluated in one language a
\emph{model--language pair}, and several analyses below are run pair by pair. Leaderboard
difficulty is the leaderboard score of \S\ref{sec:prelim}, and own-model difficulty is the score
from the other evaluated models. The \emph{own-model hard tail} is the hard tail with own-model
difficulty.

\section{Results by model and language}
\label{app:c1}
\begin{figure*}[t]\centering
\includegraphics[width=0.9\textwidth]{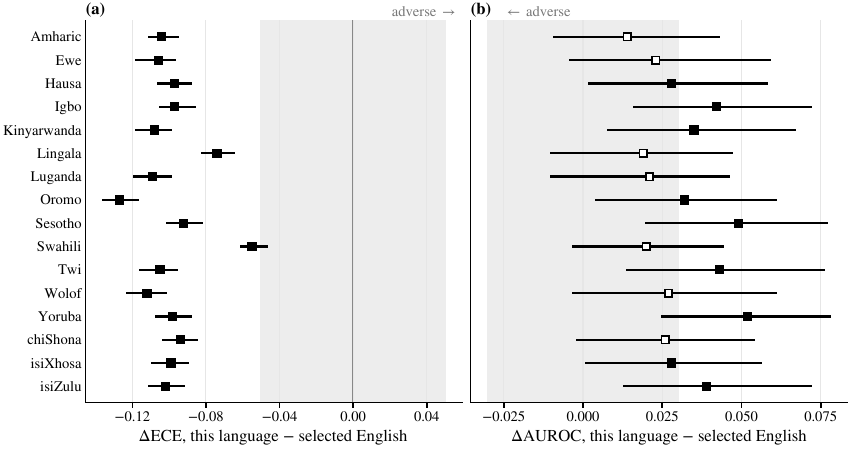}
\caption{Per-language ECE (a) and AUROC (b) differences, African minus English, under the own-model
hard tail. The ten AfriMMLU models are pooled within each language, with $95\%$ ID-clustered
bootstrap confidence intervals (CIs). Solid markers mean the CI excludes zero, and the grey bands
mark $\pm0.05$ ECE and $\pm0.03$ AUROC. Relative to this reference, African
confidence is not worse than English on either metric in any language. Compared on the same
hard-tail questions in both languages, the AUROC difference is negative (Table~\ref{tab:refset}).}
\label{fig:forest-perlang}
\end{figure*}

Table~\ref{tab:main} gives accuracy, ECE, AUROC, excess-AURC and C2MR for each of the seven models.
Fig.~\ref{fig:forest-perlang} gives the ECE and AUROC differences per language relative to the own-model hard
tail, and Table~\ref{tab:highacc} gives the same differences on the Belebele pairs with high accuracy.

In Fig.~\ref{fig:auroc-acc}a, the joint fit over all $287$ model--language pairs includes eight
English Belebele results that are not drawn, and $r=0.92$ within the African results alone.
Within the Qwen2.5 models the slope is $+0.69$, close to the $+0.68$ over all ten AfriMMLU models,
and Belebele alone gives slope $+0.55$ ($r=0.98$). The six Global-MMLU languages lie close to the
same line. The relation is mildly concave, with local slopes falling from $+0.82$ to $+0.40$ as
AUROC saturates.

\begin{table*}[t]
\centering\small
\begin{tabular}{@{}lccccccccc@{}}
\toprule
 & \multicolumn{2}{c}{Accuracy} & \multicolumn{2}{c}{ECE $\downarrow$} & \multicolumn{2}{c}{AUROC $\uparrow$} & \multicolumn{2}{c}{excess-AURC $\downarrow$} & C2MR $\uparrow$ \\
\cmidrule(lr){2-3}\cmidrule(lr){4-5}\cmidrule(lr){6-7}\cmidrule(lr){8-9}\cmidrule(l){10-10}
Model & Eng & Afr & Eng & Afr & Eng & Afr & Eng & Afr & Afr \\
\midrule
Aya-Expanse-8B & 0.55 & 0.31 & 0.41 & 0.60 & 0.728 & 0.544 & 0.14 & 0.31 & 0.04 \\
Qwen3-8B & 0.69 & 0.33 & 0.24 & 0.47 & 0.766 & 0.579 & 0.15 & 0.28 & 0.04 \\
Qwen2.5-7B & 0.69 & 0.34 & 0.27 & 0.50 & 0.818 & 0.575 & 0.06 & 0.28 & 0.05 \\
Llama-3.1-8B & 0.61 & 0.33 & 0.22 & 0.24 & 0.792 & 0.604 & 0.09 & 0.26 & 0.12 \\
Gemma-2-9B & 0.69 & 0.38 & 0.25 & 0.41 & 0.836 & 0.630 & 0.05 & 0.24 & 0.13 \\
Qwen2.5-32B & 0.82 & 0.39 & 0.15 & 0.48 & 0.895 & 0.631 & 0.02 & 0.23 & 0.07 \\
Qwen3-32B & 0.80 & 0.40 & 0.14 & 0.30 & 0.878 & 0.631 & 0.03 & 0.23 & 0.09 \\
\bottomrule
\end{tabular}
\caption{Confidence quality per model on AfriMMLU (token probability), English and the
African-language average (the two 32B models are AWQ 4-bit, added to cover high accuracy). English C2MR
is $1.0$ by definition and is not shown.}
\label{tab:main}
\end{table*}

\begin{table*}[t]
\centering\footnotesize
\begin{tabular}{@{}lccc@{}}
\toprule
Cut-off (Acc$_{\text{Afr}}\ge$) & Pairs & $\Delta$AUROC [95\% CI] & $\Delta$ECE [95\% CI] \\
\midrule
$0.50$ & $14$ & $+0.111$ $[+0.060,\allowbreak +0.171]$ & $-0.028$ $[-0.036,\allowbreak -0.021]$ \\
$0.55$ & $11$ & $+0.110$ $[+0.055,\allowbreak +0.167]$ & $-0.028$ $[-0.036,\allowbreak -0.020]$ \\
$0.60$ & $7$  & $+0.096$ $[+0.046,\allowbreak +0.149]$ & $-0.019$ $[-0.026,\allowbreak -0.012]$ \\
$0.65$ & $4$  & $+0.067$ $[+0.017,\allowbreak +0.121]$ & $-0.011$ $[-0.017,\allowbreak -0.005]$ \\
\bottomrule
\end{tabular}
\caption{High-accuracy matched test under the own-model hard tail, on the Belebele pairs of all ten models, including the two 32B models. At each cut-off we keep the pairs whose African accuracy clears it, select an English hard tail at each pair's African accuracy, and pool the per-pair differences across models and African languages. At every cut-off $\Delta\mathrm{AUROC}>0$ and
$\Delta\mathrm{ECE}<0$, and the lower CI bound of $\Delta\mathrm{AUROC}$ stays strictly
positive. The selected English questions are those on which English confidence ranks poorly
(\S\ref{sec:c1}). $\Delta$ is African minus selected English, with an ID-clustered bootstrap
($B{=}2000$) and the selection re-run in each replicate.}
\label{tab:highacc}
\end{table*}

\begin{figure}[t]\centering
\includegraphics[width=\columnwidth]{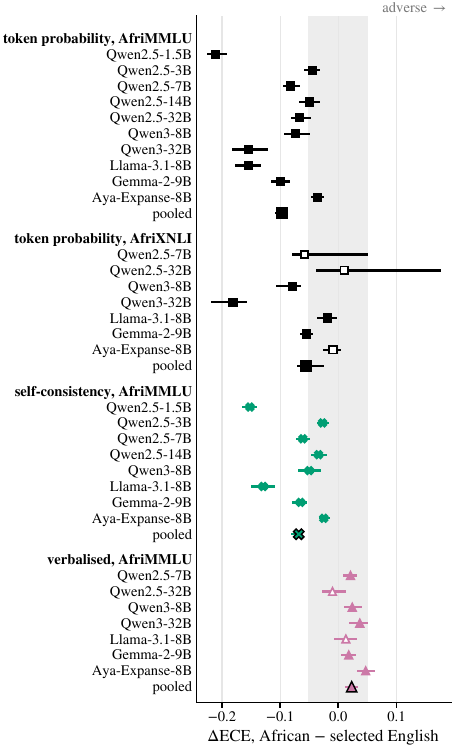}
\caption{Calibration differences per model under the own-model hard tail. Each point is $\Delta$ECE, African minus selected
English, with a $95\%$ ID-clustered bootstrap interval. Solid markers mean the interval
excludes zero, and the grey band is the $\pm0.05$ margin. The blocks use different model sets: ten
models for token probability on AfriMMLU, seven for AfriXNLI, eight (without the 32B models)
for self-consistency, and seven for verbalised confidence. Most points lie left of zero because the hard
tail keeps English questions that the models answer confidently and incorrectly, which makes
English calibration look worse. The per-language version is Fig.~\ref{fig:forest-perlang}.}
\label{fig:forest}
\end{figure}

\begin{figure}[t]\centering
\includegraphics[width=\columnwidth]{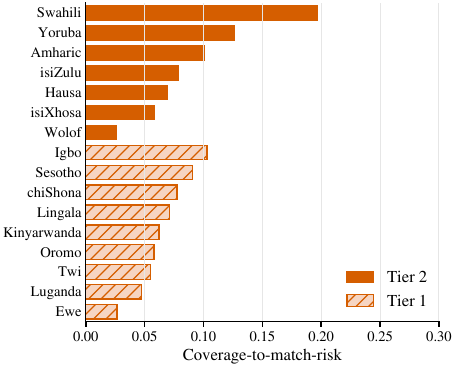}
\caption{Coverage-to-match-risk by language for token probability, averaged across the seven
models, with bars shaded and hatched by resource tier. Every African language is far
below English, with Swahili highest at $\approx 0.20$ and most languages below $0.1$ (mean over
languages $0.08$, pooled over languages $0.076$). C2MR is associated with accuracy across the 16
languages ($R^2{=}0.89$) and more weakly across model--language pairs ($R^2{=}0.49$). Coverage follows each language's accuracy more closely than its
resource tier, so some tier-2 languages fall below tier-1 languages with higher accuracy.}
\label{fig:c2mr}
\end{figure}

\begin{table}[t]
\centering\small\setlength{\tabcolsep}{3.5pt}
\begin{tabular}{@{}llcccc@{}}
\toprule
Task & Group & ECE & ECE (Platt) & AUROC & C2MR \\
\midrule
AfriMMLU & English & 0.23 & 0.07 & 0.81 & 1.000 \\
         & African & 0.43 & 0.05 & 0.59 & 0.075 \\
\midrule
AfriXNLI & English & 0.16 & 0.05 & 0.80 & 1.000 \\
         & African & 0.46 & 0.06 & 0.58 & 0.043 \\
\bottomrule
\end{tabular}
\caption{Platt recalibration for each model and language, fitted and evaluated on halves split
by ID parity (token probability, seven models). ECE and AUROC are on the evaluation halves. The
Platt map is fitted on the logit of confidence with its slope constrained to be positive, so it
cannot reorder confidence. AUROC is unchanged to four decimals, and coverage-to-match-risk is unchanged in
every model--language pair. English C2MR is $1.0$ by definition.}
\label{tab:recal}
\end{table}

\subsection{Coverage at English-level risk and recalibration}
\label{app:c3}
C2MR depends on the confidence ranking as well as on accuracy. To separate the two, we
keep correctness fixed, replace only the score used to order the African rows, and keep the
same English full-coverage risk budget. The observed ordering gives pooled C2MR $0.076$ ($95\%$
CI $[0.051,\allowbreak 0.104]$). A fixed random ordering gives $0.000$ ($95\%$ CI
$[0.000,\allowbreak 0.000]$), the paired English score gives $0.002$ $[0.000,\allowbreak 0.010]$, and ordering by
correctness gives $0.508$ $[0.483,\allowbreak 0.535]$. That last value is analytically
$a_{\mathrm{Afr}}/a_{\mathrm{Eng}}$ per model, an upper bound for these answers that only an ordering
which knows the correct answers can reach (seven models, $479$ shared question IDs, $B{=}5000$).

\section{English reference sets and difficulty scores}
\label{app:refset}

The ten models give the same pattern as the seven models of the main text (Table~\ref{tab:refset}, AfriMMLU with token probability). With either difficulty score, the matched-accuracy differences stay close to zero and the equal-difficulty differences lie between $-0.106$ and $-0.192$.

Leaderboard difficulty comes from the metabench data
\citep{kipnisMetabenchSparseBenchmark2025}, which record whether each of $5{,}226$ models from the
first Open LLM Leaderboard ($5{,}225$ for geography) answered each MMLU test question
correctly. We matched all $500$ AfriMMLU English questions to metabench items on identical
normalised question text and identical options in order, and scored each question by its
error rate across those models. None of the ten models we evaluate, or any fine-tune of them, are among
them. Earlier releases of the same model families are included, and removing the $158$ models
built on Qwen changes the difficulty ranking very little (Spearman $0.9998$). Across the $479$ questions
shared by all ten models, this difficulty correlates with the one from the other evaluated models at
$0.80$ $[0.76,\allowbreak 0.83]$, and at only $0.48$ within global facts.

The leaderboard models predict each evaluated model's own English errors less well than the other
evaluated models do (mean Spearman $0.62$ compared with $0.74$). The harder English questions
also differ in content. Global facts makes up $20.9\%$ of the benchmark and $37.0\%$ of the
English questions above the $0.6$ difficulty cut-off. Within the leaderboard-difficulty deciles,
African accuracy is lower than English accuracy by $0.46$ to $0.56$ in the easiest four deciles
and by $0.01$ in the hardest. Where the two accuracies are within ten points, English accuracy
is $0.16$ to $0.40$, near the chance level of $0.25$. Per model, calibration relative to the
own-model hard tail favours the African languages for token probability and self-consistency
but not for verbalised confidence (Fig.~\ref{fig:forest}).

Own-model difficulty for each evaluated model comes from the other models' answers only. The
own-model hard tail orders questions by the other models' error rate, and the quantile curve,
deciles and reweighting use a Rasch (one-parameter logistic) item difficulty fitted to the same
answers, which ranks questions almost identically (Spearman correlation at least $0.999$). Seven
of the ten models are Qwen models, and the score inherits their errors. With the ten models, on the hardest tenth
of questions it selects, English accuracy is $0.16$, below the $0.25$ chance rate, and English
AUROC is $0.37$. On those questions English confidence tends to rank wrong answers above
correct ones. The leaderboard score gives
$0.31$ and $0.54$. Leaving both Qwen families out of the score moves the own-model hard tail from
$+0.034$ to $+0.006$ $[-0.022,\allowbreak +0.040]$, and weighting the families equally gives $+0.013$
$[-0.010,\allowbreak +0.047]$. Removing $27$ of the $479$ questions ($5.6\%$) is enough to make the hard tail
non-positive, and removing $6$ makes the reweighted difference positive. With the same models and
difficulty, the equal-difficulty difference ($-0.151$) needs $175$ removals ($36.5\%$) to reach
zero. Leaving out any one subject keeps the hard tail between $+0.030$ and $+0.041$.

The full-benchmark gap changes little under several checks on the ten models. Leaving out any
one subject gives a gap between $-0.196$ and $-0.251$. Removing the $17$ questions that MMLU-Redux \citep{gemaAreWeDone2025}
flags among the $298$ it covers gives $-0.221$, and changes the hard tail by $-0.005$
$[-0.012,\allowbreak +0.006]$. Pooling the languages row by row, averaging per-language differences, and
comparing within each language give $-0.219$, $-0.222$ and $-0.221$. In a Murphy decomposition
\citep{murphyNewVectorPartition1973} of the binary Brier score of the top-label confidence, over
the $15$ equal-mass ECE bins, the share of uncertainty resolved by African confidence is $0.231$ lower
than by English confidence on the full benchmark and $0.011$ higher when English is restricted to the hard tail.

Across the seven models, the per-model equal-weight difference ranges from $-0.115$ (Aya-Expanse-8B) to $-0.266$
(Qwen2.5-32B) with leaderboard difficulty, and from $-0.081$ to $-0.256$ for the same two models
with own-model difficulty. Every per-model interval lies below zero. Weighting each side
by its own correct--incorrect pairs gives $-0.127$ with leaderboard difficulty and $-0.071$ with
own-model difficulty. Because this weighting compares different mixtures of deciles, the ranges in the main text leave it out. Deciles have equal size, with tied difficulties split at random, and are held fixed across
bootstrap replicates. Deciles in which a model answers every English question
correctly have no English AUROC and are dropped. With leaderboard difficulty this removes $0.14$
deciles per model on average. Removal counts come from a greedy search that at each step drops
the question whose removal moves the pooled estimate furthest towards zero, keeping the decile
boundaries of the full question set. For the reweighting, questions are instead removed in the
order of their initial influence. The reweighting tilts English
weights in proportion to $\exp(\lambda d)$ for standardised difficulty $d$, renormalised within
subjects to the benchmark shares, with $\lambda$ solved for each model in every replicate.
With leaderboard difficulty it reaches the African accuracy in $1{,}317$ of $2{,}000$ replicates
(seven models), and intervals are computed over the feasible replicates. The tilt concentrates
weight on few questions for the stronger models (effective sample size $7$--$8\%$ of the English
questions for the 32B models). The hard-tail calibration intervals are skewed because the
stopping point changes between replicates.

\begin{table*}[t]
\centering\footnotesize
\setlength{\tabcolsep}{3pt}
\resizebox{\textwidth}{!}{%
\begin{tabular}{@{}lcccc@{}}
\toprule
 & \multicolumn{2}{c}{Seven models} & \multicolumn{2}{c}{Ten models} \\
\cmidrule(lr){2-3}\cmidrule(l){4-5}
Method & Leaderboard & Other models & Leaderboard & Other models \\
\midrule
Full benchmark & $-0.216$ $[-0.233,\allowbreak -0.198]$ & $-0.216$ $[-0.233,\allowbreak -0.198]$ & $-0.219$ $[-0.236,\allowbreak -0.202]$ & $-0.219$ $[-0.236,\allowbreak -0.202]$ \\
Harder half ($q=0.5$) & $-0.084$ $[-0.111,\allowbreak -0.057]$ & $-0.059$ $[-0.090,\allowbreak -0.027]$ & $-0.081$ $[-0.108,\allowbreak -0.056]$ & $-0.039$ $[-0.072,\allowbreak -0.010]$ \\
Hardest tenth ($q=0.9$) & $+0.060$ $[-0.013,\allowbreak +0.137]$ & $+0.202$ $[+0.114,\allowbreak +0.269]$ & $+0.059$ $[-0.005,\allowbreak +0.131]$ & $+0.224$ $[+0.155,\allowbreak +0.272]$ \\
\quad English AUROC there & $0.54$ & $0.40$ & $0.54$ & $0.37$ \\
Hard tail & $+0.015$ $[-0.039,\allowbreak +0.056]$ & $+0.024$ $[0.000,\allowbreak +0.056]$ & $+0.014$ $[-0.034,\allowbreak +0.055]$ & $+0.034$ $[+0.009,\allowbreak +0.060]$ \\
Hard tail, same questions & $-0.046$ $[-0.101,\allowbreak -0.001]$ & $-0.062$ $[-0.091,\allowbreak -0.029]$ & $-0.040$ $[-0.092,\allowbreak +0.005]$ & $-0.053$ $[-0.080,\allowbreak -0.025]$ \\
Reweighting & $-0.001$ $[-0.048,\allowbreak +0.057]$ & $-0.048$ $[-0.079,\allowbreak -0.003]$ & $-0.009$ $[-0.049,\allowbreak +0.037]$ & $-0.043$ $[-0.068,\allowbreak -0.005]$ \\
Equal difficulty, equal weights & $-0.191$ $[-0.209,\allowbreak -0.172]$ & $-0.157$ $[-0.173,\allowbreak -0.137]$ & $-0.187$ $[-0.205,\allowbreak -0.168]$ & $-0.151$ $[-0.164,\allowbreak -0.130]$ \\
Equal difficulty, English pairs & $-0.158$ $[-0.183,\allowbreak -0.134]$ & $-0.116$ $[-0.140,\allowbreak -0.094]$ & $-0.157$ $[-0.182,\allowbreak -0.134]$ & $-0.106$ $[-0.129,\allowbreak -0.083]$ \\
Equal difficulty, African pairs & $-0.196$ $[-0.217,\allowbreak -0.171]$ & $-0.168$ $[-0.189,\allowbreak -0.143]$ & $-0.192$ $[-0.212,\allowbreak -0.169]$ & $-0.163$ $[-0.183,\allowbreak -0.134]$ \\
Hardest fifth, same questions & $-0.063$ $[-0.106,\allowbreak -0.018]$ & $+0.002$ $[-0.040,\allowbreak +0.043]$ & $-0.054$ $[-0.096,\allowbreak -0.010]$ & $+0.008$ $[-0.032,\allowbreak +0.051]$ \\
\bottomrule
\end{tabular}}
\caption{African minus English AUROC on AfriMMLU (token probability) under each choice of
English reference. ``Leaderboard'' scores English difficulty by the error rate of $5{,}226$
metabench models, and ``other models'' by the errors of the other evaluated models. Intervals are
$95\%$ ID-clustered bootstrap intervals over $2{,}000$ replicates. The reweighting is feasible in
$1{,}317$ (seven models) and $1{,}316$ (ten models) of them with leaderboard difficulty, and in
$1{,}935$ and $1{,}965$ with the other models' difficulty.}
\label{tab:refset}
\end{table*}

\subsection{Results above accuracy thresholds and for larger models}
\label{app:verb-floors}

On AfriMMLU the full-question difference persists at every accuracy threshold, while every own-model hard-tail AUROC
interval for the seven models includes zero (Table~\ref{tab:verb-floors},
which also gives subsets of stronger models). Only two of their pairs meet the 0.50 accuracy
threshold. On AfriXNLI, positive hard-tail differences remain in the higher-accuracy subsets, but in
some pairs the selected English accuracy is far from the African accuracy, by up to 22.7 points
with the held-out cut-off over all pairs.

We use five accuracy thresholds. The first keeps all pairs, and the others keep pairs with
accuracy strictly above chance, strictly above chance plus 0.05 or 0.10, and at least 0.50. Chance is $1/4$ for AfriMMLU and $1/3$ for AfriXNLI. Eligibility
is fixed from observed common-ID accuracies before resampling. Consequently,
intervals condition on the selected pairs and do not incorporate uncertainty in
whether a pair would qualify on new questions.

This audit uses the seven models and 112 model--African-language pairs. On AfriMMLU, 477
question IDs are common to every pair (600 on AfriXNLI), two fewer than the 479 shared
by every model in English that the analyses above use. Its control is the own-model
hard tail, matched to each model and African language separately, and the reported average weights
qualifying pairs equally. The \emph{adaptive} control chooses the hard-tail cut-off on the same
outcomes it is scored on. The \emph{held-out} control chooses it on four of five question-ID folds
and applies it to the fifth (fold seed $20260915$). The population and the quantity estimated both differ from those of the pooled
analyses in the main text.

Among the seven models, only the two quantised 32B models exceed 14B, so the $\geq14$B and
32B subsets coincide. A separate AfriMMLU analysis adds Qwen2.5-14B, which also changes the set
of models that scores difficulty, giving eight models of at least 7B and three of at least 14B
on 477 common question IDs. The 32B held-out difference is $-0.007$ in that analysis and $+0.010$ in the seven-model one (Table~\ref{tab:verb-floors}), and we do not pool the two. For the stronger models, neither analysis gives a precise
estimate of the AfriMMLU hard-tail difference. The high-accuracy test of Table~\ref{tab:highacc} uses Belebele pairs only.

\begin{table*}[t]
\centering\footnotesize
\setlength{\tabcolsep}{2pt}
\begin{tabular}{@{}llrcccrr@{}}
\toprule
Models & Accuracy rule & $N$ & Raw $\Delta$AUROC & Adaptive $\Delta$AUROC & Held-out $\Delta$AUROC & Gap: mean / max & Valid $B$ \\
\midrule
\multicolumn{8}{l}{\textit{AfriMMLU: seven models, 477 common IDs}} \\
Seven & All & 112 & \estci{-.220}{-.239}{-.201} & \estci{+.021}{-.005}{+.050} & \estci{+.024}{-.007}{+.049} & $+0.2\,/\,2.2$  & 1997 \\
Seven & $a>c$ & 112 & \estci{-.220}{-.239}{-.201} & \estci{+.021}{-.005}{+.050} & \estci{+.024}{-.007}{+.049} & $+0.2\,/\,2.2$  & 1997 \\
Seven & $a>c+.05$ & 101 & \estci{-.221}{-.239}{-.202} & \estci{+.020}{-.007}{+.051} & \estci{+.022}{-.009}{+.048} & $+0.2\,/\,2.2$  & 1998 \\
Seven & $a>c+.10$ & 50 & \estci{-.227}{-.247}{-.206} & \estci{+.020}{-.018}{+.062} & \estci{+.023}{-.022}{+.058} & $+0.2\,/\,1.9$  & 2000 \\
Seven & $a\geq.50$ & 2 & \estci{-.154}{-.181}{-.126} & \estci{+.011}{-.030}{+.059} & \estci{+.017}{-.030}{+.060} & $-0.5\,/\,0.6$  & 2000 \\
32B & All & 32 & \estci{-.259}{-.284}{-.235} & \estci{+.003}{-.055}{+.072} & \estci{+.010}{-.060}{+.069} & $+0.2\,/\,1.9$  & 1999 \\
32B & $a\geq.50$ & 1 & \estci{-.179}{-.215}{-.143} & \estci{+.025}{-.032}{+.102} & \estci{+.032}{-.034}{+.100} & $-0.6\,/\,0.6$  & 2000 \\
\midrule
\multicolumn{8}{l}{\textit{AfriXNLI: seven models, 600 common IDs}} \\
Seven & All & 112 & \estci{-.206}{-.230}{-.182} & \estci{+.094}{+.049}{+.150} & \estci{+.084}{+.035}{+.129} & $+3.2\,/\,22.7$  & 1963 \\
Seven & $a>c$ & 109 & \estci{-.204}{-.228}{-.181} & \estci{+.094}{+.050}{+.149} & \estci{+.085}{+.037}{+.130} & $+3.0\,/\,18.7$  & 1965 \\
Seven & $a>c+.05$ & 71 & \estci{-.205}{-.232}{-.178} & \estci{+.127}{+.070}{+.182} & \estci{+.114}{+.054}{+.164} & $+3.6\,/\,15.9$  & 1967 \\
Seven & $a>c+.10$ & 40 & \estci{-.183}{-.212}{-.154} & \estci{+.155}{+.095}{+.210} & \estci{+.142}{+.080}{+.194} & $+3.0\,/\,10.1$  & 1970 \\
Seven & $a\geq.50$ & 15 & \estci{-.153}{-.181}{-.124} & \estci{+.143}{+.092}{+.202} & \estci{+.139}{+.087}{+.192} & $+1.7\,/\,5.0$  & 1987 \\
32B & All & 32 & \estci{-.215}{-.256}{-.169} & \estci{+.201}{+.100}{+.288} & \estci{+.178}{+.077}{+.263} & $+7.0\,/\,22.7$  & 1967 \\
32B & $a\geq.50$ & 8 & \estci{-.176}{-.216}{-.133} & \estci{+.198}{+.112}{+.297} & \estci{+.194}{+.103}{+.279} & $+2.6\,/\,5.0$  & 1987 \\
\midrule
\multicolumn{8}{l}{\textit{Secondary AfriMMLU analysis: difficulty also scored with Qwen2.5-14B, 477 IDs}} \\
$\geq7$B & All & 128 & \estci{-.221}{-.239}{-.203} & \estci{+.034}{+.006}{+.062} & \estci{+.036}{+.005}{+.061} & $+0.2\,/\,3.0$  & 1999 \\
$\geq14$B & All & 48 & \estci{-.250}{-.274}{-.226} & \estci{+.035}{-.019}{+.084} & \estci{+.034}{-.021}{+.082} & $+0.1\,/\,1.6$  & 1999 \\
$\geq14$B & $a\geq.50$ & 1 & \estci{-.179}{-.215}{-.143} & \estci{+.031}{-.031}{+.094} & \estci{+.037}{-.033}{+.092} & $-0.1\,/\,0.1$  & 2000 \\
32B & All & 32 & \estci{-.259}{-.284}{-.235} & \estci{-.004}{-.066}{+.056} & \estci{-.007}{-.070}{+.052} & $+0.1\,/\,1.6$  & 1999 \\
\bottomrule
\end{tabular}
\caption{Sensitivity to the accuracy threshold, on saved token-probability outputs.
$a$ is observed African answer accuracy and $c$ is task chance. All five
accuracy thresholds are shown for the seven models. $N$ counts
qualifying model--language pairs. Differences are African minus English AUROC.
Brackets contain 95\% intervals conditional on fixed eligibility and estimability.
The final columns report held-out English-minus-African mean accuracy gap and
maximum absolute per-pair gap, in percentage points, and the number of valid held-out replicates. Adaptive rows retain all 2000 replicates. Replicates in which the held-out difference cannot be
computed for some pair are dropped, which gives the counts below 2000. On AfriMMLU the ``All''
and $a>c$ rows coincide because every one of the 112 pairs is above chance.}
\label{tab:verb-floors}
\end{table*}

\section{Audits of the verbalised comparison}
\label{app:verb}

\subsection{Each prompt on its own answers}
\label{app:c2}
\begin{figure*}[t]\centering
\includegraphics[width=\textwidth]{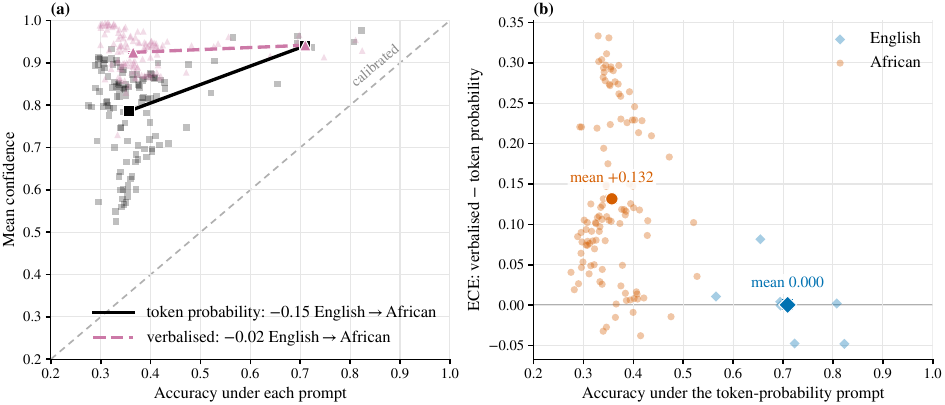}
\caption{Calibration of each prompt evaluated on its own answers, over
questions where both responses parse. \textbf{(a)}
Means for each model and language, and English and African group means. \textbf{(b)} ECE, verbalised minus
token probability, which is $+0.132$ $[+0.126,\allowbreak +0.137]$ in the African languages and $+0.000$
$[-0.005,\allowbreak +0.007]$ in English. Intervals resample shared question IDs across fixed models,
languages, option orders and prompts ($B=2{,}000$).}
\label{fig:saysknow}
\end{figure*}

Fig.~\ref{fig:saysknow} shows each prompt's confidence and the accuracy of its own answers for the seven models. Evaluating both prompts on token-probability correctness would be invalid where the two prompts give different answers. On AfriMMLU rows where both prompts parse, ECE is $0.561$ (verbalised) and $0.429$ (token probability) in the African languages and $0.231$ for both in English. By model, the African difference runs from $-0.002$ (Qwen2.5-32B) to $+0.286$ (Llama-3.1-8B), and the English difference from $-0.048$ to $+0.082$. The two prompts choose the same option on $44.6\%$ to $90.0\%$ of African AfriMMLU questions, depending on the model (Appendix~\ref{app:verb-prompts}). On the same rows, AUROC is $0.68$ (verbalised) and $0.83$ (token probability) in English and $0.55$ and $0.60$ in the African languages. With leaderboard difficulty, verbalised calibration still favours English when English is restricted to the hard tail ($\Delta\mathrm{ECE}=+0.030$ $[+0.014,\allowbreak +0.078]$). There, token-probability calibration favours the African languages ($-0.094$).

\subsection{Answer agreement and parsing}
\label{app:verb-prompts}

Answer agreement is only 73.0\% across African AfriMMLU model--language pairs and 79.1\% across
African AfriXNLI pairs (Table~\ref{tab:verb-prompts}). Similar aggregate
accuracies therefore hide substantial changes in the predicted answers. On
AfriMMLU, conditioning on both prompts parsing also reverses the aggregate accuracy
comparison. Verbalised (VB) accuracy exceeds token-probability (TP) accuracy by 0.82 points on
paired examples, whereas VB accuracy over all prompts is 33.4\%, below TP's 35.4\%.

We paired TP and VB records by model, task,
language, question ID and option order, and checked that the gold labels agree. The saved predictions give the
predicted option and the number of prompts for all seven models on both
benchmarks.

\begin{table*}[t]
\centering\footnotesize
\setlength{\tabcolsep}{4pt}
\begin{tabular}{@{}llcccc@{}}
\toprule
Task & Group & TP parse rate & VB parse rate & TP all-prompt accuracy & VB all-prompt accuracy \\
\midrule
AfriMMLU & English & \estci{.992}{.989}{.994} & \estci{.927}{.914}{.939} & \estci{.689}{.664}{.715} & \estci{.659}{.633}{.685} \\
& African & \estci{.9997}{.9996}{.9999} & \estci{.911}{.902}{.919} & \estci{.354}{.336}{.370} & \estci{.334}{.317}{.350} \\
AfriXNLI & English & \estci{1.000}{1.000}{1.000} & \estci{.999}{.998}{.999} & \estci{.787}{.768}{.805} & \estci{.774}{.757}{.792} \\
& African & \estci{1.000}{1.000}{1.000} & \estci{.978}{.977}{.979} & \estci{.421}{.406}{.436} & \estci{.413}{.400}{.425} \\
\midrule
Task & Group & TP paired accuracy & VB paired accuracy & Own-answer $\Delta$ECE & Answer agreement \\
\midrule
AfriMMLU & English & $.709$ & $.710$ & \estci{+.000}{-.005}{+.007} & \estci{.897}{.887}{.907} \\
& African & $.357$ & $.365$ & \estci{+.132}{+.126}{+.137} & \estci{.730}{.723}{.737} \\
AfriXNLI & English & $.787$ & $.775$ & \estci{-.014}{-.022}{+.007} & \estci{.917}{.910}{.924} \\
& African & $.422$ & $.421$ & \estci{+.009}{+.004}{+.013} & \estci{.791}{.787}{.795} \\
\bottomrule
\end{tabular}
\caption{Own-answer prompt audit for the seven models on each task. Entries
are equal-model, then equal-language means, with 95\% paired question-ID bootstrap
intervals. The upper half gives parse rates and accuracy over every prompted example, counting
unparseable answers as wrong, and the lower half gives metrics on examples parseable under both
prompts. Agreement counts identical predicted options, which is stricter than identical
correctness. Paired accuracies are point estimates. $\Delta$ECE is VB minus TP,
using each prompt's own answers and 15 adaptive bins. The two halves have different
denominators and should be read together.}
\label{tab:verb-prompts}
\end{table*}

On the questions where both prompts give the same answer, the verbalised minus token-probability ECE in the African languages is $+0.1052$ $[0.1005,\allowbreak 0.1091]$ on
AfriMMLU, but $-0.0210$ $[-0.0233,\allowbreak -0.0196]$ on AfriXNLI, compared with $+0.0091$
$[0.0045,\allowbreak 0.0133]$ on all paired AfriXNLI examples. This subset fixes correctness
across prompts by selection and may change the question population substantially,
so it does not isolate a causal effect of how confidence is elicited.

\subsection{Failed parses}
\label{app:parse}
The verbalised result of \S\ref{sec:c2} conditions on a parseable answer and confidence.
Table~\ref{tab:parse-rates} gives parse rates per language. 

Counting failed verbalised responses as wrong, with either the model's mean confidence in that language or unit confidence, raises pooled per-language ECE on both sides, by $0.029$ and
$0.039$ on average in the African languages and by $0.051$ and $0.056$ in English.
These imputations do not observe the missing confidence and do not bound the matched verbalised result. We did not rerun the matched comparison with them, because doing so would change the target accuracies, the selected English items and the ECE bins. Whether the matched estimate depends on how failed responses are handled is therefore untested.

\begin{table}[t]
\centering\small
\begin{tabular}{lcc@{\hskip 1em}lcc}
\toprule
lang & parse & $\mathrm{acc}^{\mathrm{tp}}_{\mathrm{unp.}}$ &
lang & parse & $\mathrm{acc}^{\mathrm{tp}}_{\mathrm{unp.}}$ \\
\midrule
eng & .93 & .48 & xho & .91 & .29 \\
ibo & .96 & .31 & kin & .91 & .30 \\
swa & .95 & .31 & twi & .91 & .29 \\
yor & .95 & .33 & amh & .90 & .30 \\
sot & .93 & .29 & sna & .90 & .31 \\
hau & .93 & .32 & lin & .89 & .31 \\
orm & .92 & .31 & ewe & .85 & .29 \\
zul & .91 & .29 & wol & .84 & .29 \\
lug & .91 & .29 &  &  &  \\
\bottomrule
\end{tabular}
\caption{Verbalised parse rate per language (ISO 639-3 codes, pooled over the seven models,
AfriMMLU) and token-probability accuracy of the unparsed items. Token-probability
accuracy on parsed items is $0.36$ across the African languages (mean over languages) and $0.71$
in English.}
\label{tab:parse-rates}
\end{table}

\section{Setup and reproducibility}
\label{app:repro}

\subsection{Experimental setup}
\label{app:setup}
Table~\ref{tab:langs} lists the $16$ African languages and English by Joshi resource tier.
Table~\ref{tab:roster} lists the models we evaluate, and Table~\ref{tab:signal-taxonomy} gives the
three confidence signals with their access requirements.

\begin{table}[t]
\centering\small
\begin{tabular}{@{}lp{0.72\columnwidth}@{}}
\toprule
Tier & Languages \\
\midrule
5 (baseline) & English \\
2 & Swahili, Hausa, Yoruba, Amharic, isiZulu, isiXhosa, Wolof \\
1 & Igbo, chiShona, Kinyarwanda, Sesotho, Luganda, Lingala, Oromo, Twi, Ewe \\
\bottomrule
\end{tabular}
\caption{The 16 African languages and English of AfriMMLU, by Joshi resource
tier \citep{joshiStateFateLinguistic2020}. English is the within-item baseline.}
\label{tab:langs}
\end{table}

\begin{table*}[t]
\centering\footnotesize\setlength{\tabcolsep}{3pt}
\begin{tabular}{@{}lllc@{}}
\toprule
Model & Hugging Face repository & Note & Params \\
\midrule
\multicolumn{4}{@{}l}{\emph{In the analyses (ten models)}}\\
Aya-Expanse-8B$^{\dagger}$ & \texttt{CohereLabs/aya-expanse-8b} & multilingual-first & 8B \\
Llama-3.1-8B$^{\dagger}$ & \texttt{meta-llama/Llama-3.1-8B-Instruct} & & 8B \\
Gemma-2-9B$^{\dagger}$ & \texttt{google/gemma-2-9b-it} & & 9B \\
Qwen3-8B$^{\dagger}$ & \texttt{Qwen/Qwen3-8B} & & 8B \\
Qwen3-32B-AWQ$^{\dagger}$ & \texttt{Qwen/Qwen3-32B-AWQ} & high accuracy & 32B (4-bit) \\
Qwen2.5-1.5B & \texttt{Qwen/Qwen2.5-1.5B-Instruct} & size series & 1.5B \\
Qwen2.5-3B & \texttt{Qwen/Qwen2.5-3B-Instruct} & size series & 3B \\
Qwen2.5-7B$^{\dagger}$ & \texttt{Qwen/Qwen2.5-7B-Instruct} & size series & 7B \\
Qwen2.5-14B & \texttt{Qwen/Qwen2.5-14B-Instruct} & size series & 14B \\
Qwen2.5-32B-AWQ$^{\dagger}$ & \texttt{Qwen/Qwen2.5-32B-Instruct-AWQ} & size series & 32B (4-bit) \\
\midrule
\multicolumn{4}{@{}l}{\emph{African-tuned (at or below chance, not analysed)}}\\
InkubaLM-0.4B & \texttt{lelapa/InkubaLM-0.4B} & small, for limited compute & 0.4B \\
Lugha-Llama-8B-wura & \texttt{Lugha-Llama/Lugha-Llama-8B-wura} & continued pretraining of Llama-3.1 & 8B \\
AfroLlama-V1 & \texttt{Jacaranda/AfroLlama\_V1} & low-rank adaptation of Llama-3 & 8B \\
\bottomrule
\end{tabular}
\caption{The models we evaluate, by requested Hugging Face repository. $^{\dagger}$Used in the
seven-model analyses. The African-tuned models score
at or below chance even in English, so they are left out of the analyses (\S\ref{sec:setup}). The models are described in their
reports and model cards \citep{dangAyaExpanseCombining2024,tonjaInkubaLMSmallLanguage2024,buzaabaLughaLlamaAdaptingLarge2025,jacarandaAfroLlamaV1,qwenteamQwen25Technical2024,qwenteamQwen3Technical2025,gemmateamGemma2Improving2024,llamateamLlama3Herd2024}.}
\label{tab:roster}
\label{tab:repro-models}
\end{table*}

\begin{table}[t]
\centering\small
\begin{tabular}{@{}lp{0.34\columnwidth}cc@{}}
\toprule
Signal & Confidence score & Logits & APIs \\
\midrule
Token probability & normalised logprob over option letters & yes & no \\
Verbalised & a probability stated in words & no & yes \\
Self-consistency & agreement across $n{=}10$ samples & no & yes \\
\bottomrule
\end{tabular}
\caption{The three confidence signals. ``Logits'' marks whether a signal needs logit access,
and ``APIs'' whether it applies to closed APIs. Verbalised confidence and self-consistency
apply to any text-output model.}
\label{tab:signal-taxonomy}
\end{table}

\subsection{Reproducibility details}
\label{app:repro-details}
The repository at \url{https://github.com/sgrimbly/africal-calibration} contains the evaluation
and analysis code, each model's confidence and correctness on every question, the run records
with the generated replies, and instructions to
reproduce every table and figure. Inference uses an offline, batched vLLM backend
\citep{kwonEfficientMemoryManagement2023}. Exact model revisions were recorded for all but the
earliest runs and are in the run records released with the code. Total inference compute was on the order of
$100$ GPU-hours on consumer RTX 4090 GPUs.

\paragraph{Reference-set analyses.} The analyses of \S\ref{sec:c1} and
Appendix~\ref{app:refset} use only the saved predictions, with leaderboard difficulty taken from
the metabench data in Zenodo record 12819251. All use $2{,}000$
bootstrap replicates with seed $20261106$. The hard tail adds English questions from hardest to
easiest and stops at the first question that brings pooled English accuracy to at least the
African accuracy, provided more than $100$ English rows are kept. The reweighting solves for the
tilt that makes weighted English accuracy equal to the African accuracy, searching between
$\lambda=-30$ and $+30$ when the accuracy differences at the two ends have opposite signs.
Replicates without such a bracket are left out, and Table~\ref{tab:c1} gives how many remain.
Weighted accuracy need not change monotonically with the tilt, so some excluded replicates may
still have a matching tilt.
Own-model difficulty for each evaluated model is computed from the other models' answers only.

\paragraph{Models.} Table~\ref{tab:repro-models} lists the models we evaluate by Hugging Face
repository. The ten models that score above chance enter the analysis. We use a model's
African-language results only if it scores above multiple-choice chance in English and within
the range expected for its size on English MMLU. This checks our evaluation code. It does not
reproduce a specific published value, because per-model English values for these models are
published mainly in figures or leaderboards. Aya-Expanse-8B, the pilot model, scores $0.560$.
Reasoning models (Qwen3) are run with
\texttt{enable\_thinking=False} to suppress the \texttt{<think>} preamble. Models up to 14B run in
\texttt{bfloat16}. The two 32B checkpoints are the official AWQ 4-bit quantisations.

\paragraph{Tasks and data.} Table~\ref{tab:repro-tasks} lists the benchmarks. Each question is asked
once for every cyclic shift of its answer options to control for position bias, four times on the
four-option tasks and three times on AfriXNLI. The runs also included a fourth AfriXNLI shift, which
on three options repeats the first. It is excluded from the analysis so that no option order is
double-weighted. AfriMMLU and AfriXNLI are the IrokoBench
components \citep{adelaniIrokoBenchNewBenchmark2024}. AfriXNLI is presented as a three-option multiple-choice task with
label order $0{=}$entailment, $1{=}$neutral, $2{=}$contradiction. Lingala AfriXNLI is at chance
for every system in \citet{adelaniIrokoBenchNewBenchmark2024} and for both 32B models we evaluate, which
suggests faulty items. Leaving it out changes the seven-model AfriXNLI gap from $-0.206$ to
$-0.199$.

\begin{table*}[t]
\centering\small
\begin{tabular}{@{}llccc@{}}
\toprule
Task & Hugging Face repository & Items/lang & Opts & Langs \\
\midrule
AfriMMLU & \texttt{masakhane/afrimmlu} & 500 & 4 & 16+en \\
AfriXNLI & \texttt{masakhane/afrixnli} & 600 & 3 & 16+en \\
Belebele & \texttt{facebook/belebele} & 900 & 4 & 14+en \\
Global-MMLU & \texttt{CohereLabs/Global-MMLU} & 2000 (sampled) & 4 & 6+6+en \\
Uhura ARC-Easy & \texttt{masakhane/uhura-arc-easy} & 300--520 & 4 & 5+en \\
MMLU (English check) & \texttt{cais/mmlu} & --- & 4 & en \\
\bottomrule
\end{tabular}
\par\smallskip
{For Global-MMLU, the first six languages are African and the second six are non-African.
Belebele is CC-BY-SA-4.0. Its questions were written in English for FLORES-200
passages and translated by humans, so it is also translated from English,
but from a source other than MMLU.}
\caption{Benchmarks. Items are per language, and all use the \texttt{test} split.}
\label{tab:repro-tasks}
\end{table*}

\paragraph{Decoding and elicitation.} Table~\ref{tab:repro-decoding} gives the sampling parameters
per signal (\texttt{max\_model\_len}${=}4096$, or $8192$ for some Llama-3.1-8B runs). The token-probability and verbalised
prompts decode greedily (temperature $0$). Self-consistency uses
temperature $0.7$ with $n{=}10$ samples ($n{=}40$ in the sampling-budget experiment, all else
fixed). With $n{=}40$, African AUROC rises by at most $0.024$ for the three models tested. The token-probability signal requests the top-$20$ first-token log-probabilities, so an
option letter outside the top-$20$ is treated as zero mass. The stored per-row option mass bounds
the effect. In the most affected case (Qwen2.5-14B, English Belebele) $65\%$ of rows lack at least
one letter, yet only $3$ of $3{,}600$ rows place less than half their first-token mass on the
captured letters. The translate-test uses the fixed
\texttt{facebook/nllb-200-distilled-600M} checkpoint.

\begin{table}[t]
\centering\small
\setlength{\tabcolsep}{2pt}%
\begin{tabular}{@{}lccl@{}}
\toprule
Signal & Temp. & \texttt{max\_tok} & Confidence \\
\midrule
Token-prob & 0.0 & 4 & letter softmax \\
Verbalised & 0.0 & 48 & stated \%\,$/100$ \\
Self-cons. & 0.7 & 4 & modal agreement, $n{=}10/40$ \\
\bottomrule
\end{tabular}
\caption{Decoding parameters and the confidence score per signal.}
\label{tab:repro-decoding}
\end{table}

\paragraph{Prompt format.} The prompt is the question, a blank line, the options as
``\texttt{A.\ \dots}'' one per line, a blank line, then the instruction for the signal:
\begin{quote}\small
\emph{Token-prob / self-consistency:} ``Respond with only the single letter of the correct
option.''\\[2pt]
\emph{Verbalised:} ``Respond with the single letter of the correct option, then your confidence that
it is correct as a percentage from 0 to 100. Format: `Letter: X, Confidence: Y'.''
\end{quote}
For token probability, the predicted label is the option letter with the highest first-token log-probability among the top $20$. For self-consistency, it is the most frequent first standalone option letter across the parsed samples. For verbalised confidence, the label is the letter after ``Letter'' (else the first standalone option letter), and the confidence is the number after ``Confidence'' (else a number directly after a leading letter, else the only number in the reply). A reply without both is counted as unparsed. The instruction
language is English throughout.

\paragraph{Statistics.} We report adaptive ECE with $15$ equal-mass bins. Confidence intervals come
from a bootstrap that resamples question IDs ($B{=}1000$--$10{,}000$). For the own-model hard tail
the English selection is re-run inside every replicate, and the grey bands in
Figs.~\ref{fig:forest} and~\ref{fig:forest-perlang} mark differences of $\pm0.03$ AUROC and
$\pm0.05$ ECE. Resampling question IDs keeps translations and option orders together but does not
account for AfriXNLI questions that share a premise or Belebele questions that share a passage.

\phantomsection\label{app:verb-statistics}
The accuracy-threshold and prompt audits (Appendices~\ref{app:verb-floors}
and~\ref{app:verb-prompts}) use saved predictions, with no new model runs. Their intervals use
$2{,}000$ question-ID bootstrap replicates, with each draw shared across the compared models and
prompts.

\end{document}